\def\year{2022}\relax
\documentclass[letterpaper]{article} 
\usepackage{aaai22}  
\usepackage{times}  
\usepackage{helvet}  
\usepackage{courier}  
\usepackage[hyphens]{url}  
\usepackage{graphicx} 
\usepackage{natbib}  
\usepackage{caption} 
\usepackage{amsmath,amssymb,amsfonts}
\usepackage{pifont}
\usepackage{xspace}
\usepackage{bbm}

\usepackage{footmisc}

\usepackage{multirow}

\usepackage{algorithm}
\usepackage[noend]{algpseudocode}

\usepackage{xcolor}
\definecolor{red}{rgb}{0.8, 0.0, 0.0}
\definecolor{blue}{rgb}{0.0, 0.0, 0.8}
\definecolor{bb}{rgb}{0.0, 0.0, 0.5}
\definecolor{deeppink}{rgb}{1.0, 0.08, 0.58}
\usepackage[colorlinks=true,linkcolor=red,citecolor=bb,urlcolor=deeppink]{hyperref}

\makeatletter
\DeclareRobustCommand\onedot{\futurelet\@let@token\@onedot}
\def\@onedot{\ifx\@let@token.\else.\null\fi\xspace}

\makeatother

\DeclareCaptionStyle{ruled}{labelfont=normalfont,labelsep=colon,strut=off} 
\usepackage{newfloat}
\usepackage{listings}
\title{AutoGCL: Automated Graph Contrastive Learning via \\ Learnable View Generators}
\author{
    Yihang Yin\textsuperscript{\rm 1}, 
    Qingzhong Wang\textsuperscript{\rm 2}, 
    Siyu Huang\textsuperscript{\rm 3}, 
    Haoyi Xiong\textsuperscript{\rm 2}, 
    Xiang Zhang\textsuperscript{\rm 4}
}
\affiliations{
    \textsuperscript{\rm 1}Nanyang Technological University, \textsuperscript{\rm 2}Baidu Research, \textsuperscript{\rm 3}Harvard University, \textsuperscript{\rm 4}The Pennsylvania State University\\
    yyin009@e.ntu.edu.sg, wangqingzhong@baidu.com, huang@seas.harvard.edu, xionghaoyi@baidu.com, xzz89@psu.edu
}

\usepackage{bibentry}

\begin{document}

\maketitle

\begin{abstract}
Contrastive learning has been widely applied to graph representation learning, where the view generators play a vital role in generating effective contrastive samples. Most of the existing contrastive learning methods employ pre-defined view generation methods, e.g., node drop or edge perturbation, which usually cannot adapt to input data or preserve the original semantic structures well. To address this issue, we propose a novel framework named \emph{\underline{Auto}mated \underline{G}raph \underline{C}ontrastive \underline{L}earning} (AutoGCL) in this paper. Specifically, AutoGCL employs a set of learnable graph view generators orchestrated by an auto augmentation strategy, where every graph view generator learns a probability distribution of graphs conditioned by the input. While the graph view generators in AutoGCL preserve the most representative structures of the original graph in generation of every contrastive sample, the auto augmentation learns policies to introduce adequate augmentation variances in the whole contrastive learning procedure. Furthermore, AutoGCL adopts a joint training strategy to train the learnable view generators, the graph encoder, and the classifier in an end-to-end manner, resulting in topological heterogeneity yet semantic similarity in the generation of contrastive samples. Extensive experiments on semi-supervised learning, unsupervised learning, and transfer learning demonstrate the superiority of our AutoGCL framework over the state-of-the-arts in graph contrastive learning. In addition, the visualization results further confirm that the learnable view generators can deliver more compact and semantically meaningful contrastive samples compared against the existing view generation methods. Our code is available at \url{https://github.com/Somedaywilldo/AutoGCL}.
\end{abstract}

\section{Introduction}

Graph neural networks (GNNs)~\cite{kipf2016gcn,velivckovic2017gat,xu2018gin,hamilton2017graphsage} are gaining increasing attention in the realm of graph representation learning. By generally following a recursive neighborhood aggregation scheme, GNNs have shown impressive representational power in various domains, such as point clouds~\cite{shi2020pointgnn}, social networks~\cite{fan2019gnnsocial}, chemical analysis~\cite{de2018molgan}, and so on. Most of the existing GNN models are trained in an end-to-end supervised fashion, which relies on a high volume of fine-annotated data. However, labeling graph data requests a huge amount of effort from professional annotators with domain knowledge. To alleviate this issue, GAE~\cite{kipf2016vgae} and GraphSAGE~\cite{hamilton2017graphsage} have been proposed to exploit a naive unsupervised pretraining strategy that reconstructs the vertex adjacency information. Some recent works~\cite{hu2019pretraingnn,you2020sslgcn} introduce self-supervised pretraining strategies to GNNs which further improve the generalization performance. 


More recently, with developments of contrastive multi-view learning in computer vision \cite{he2020moco, chen2020simclr,tian2019cmc} and natural language processing \cite{yang2019xlnet,logeswaran2018efficient}, some self-supervised pretraining approaches perform as good as (or even better than) supervised methods.
%
In general, contrastive methods generate training views using data augmentations, where views of the same (positive pairs) input are concentrated in the representation space with~
views of different inputs (negative pairs) pushed apart. 
To work on graphs, DGI \cite{velivckovic2018dgi} has been proposed to treat both graph-level and node-level representations of the same graph as positive pairs, pursuing consistent representations from local and global features. CMRLG \cite{hassani2020cmcgnn} achieves a similar goal by grouping adjacency matrix (local features) and its diffusion matrix (global features) as positive pairs. GCA \cite{zhu2020gca} generates the positive view pairs through sub-graph sampling with the structure priors with node attributes randomly masked. GraphCL~\cite{you2020graphcl} offers even more strategies for augmentations, such as node dropping and edge perturbation. While above attempts incorporate contrastive learning into graphs, they usually fail to generate views with respect to the semantic of original graphs or adapt augmentation policies to specific graph learning tasks.



Blessed by the invariance of image semantics under various transformation, image data augmentation has been widely used~\cite{cubuk2019autoaugment} to generative contrastive views. However, the use of graph data augmentation might be ineffective here, as transformations on a graph might severely disrupt its semantics and properties for learning. In the meanwhile, InfoMin~\cite{tian2020goodview} improves contrastive learning for vision tasks and proposes to replace image data augmentation with a flow-based generative model for contrastive views generation. Thus, learning a probability distribution of contrastive views conditioned by an input graph might be an alternative to simple data augmentation for graph contrastive learning but still requests non-trivial efforts, as the performance and scalability of common graph generative models are poor in real-world scenarios. 

\begin{table}[!tb]
    \caption{
    An overview of graph augmentation methods. The explanation of these properties can be found in Section \ref{sec-insight}.
    }
    \vspace{-0.3cm}
    \begin{minipage}{1\linewidth}
    \centering
    \resizebox{1\hsize}{!}{
        \begin{tabular}{cccccccc}
        \hline
        Property           & CMRLG          & GRACE         & GraphCL       & GCA           & JOAO          & AD-GCL        & Ours \\
        \hline
        Topological        & $\checkmark$   & $\checkmark$  & $\checkmark$  & $\checkmark$  & $\checkmark$  & $\checkmark$  & $\checkmark$  \\
        Node  Feature      & -              & $\checkmark$  & $\checkmark$  & $\checkmark$  & $\checkmark$  & -             & $\checkmark$  \\
        Label-preserving   & -              & -             & -             & -             & -             & -             & $\checkmark$  \\
        Adaptive           & -              & -             & -             & $\checkmark$  & $\checkmark$  & $\checkmark$  & $\checkmark$  \\
        Variance           & -              & $\checkmark$  & $\checkmark$  & $\checkmark$  & $\checkmark$  & $\checkmark$  & $\checkmark$  \\
        Differentiable     & -              & -             & -             & -             & -             & $\checkmark$  & $\checkmark$  \\
        Efficient   BP     & -              & -             & -             & -             & -             & $\checkmark$  & $\checkmark$  \\
        \hline
        \end{tabular}
    }
    \end{minipage}
\label{tab-aug-compare}
\vspace{-0.6cm}
\end{table}

In this work, we propose a learnable graph view generation method, namely AutoGCL, to address above issues via learning a probability distribution over node-level augmentations. While the conventional pre-defined view generation methods, such as random dropout or graph node masking, may inevitably change the semantic labels of graphs and finally hurt contrastive learning, AutoGCL adapts to the input graph such that it can well preserve the semantic labels of the graph. In addition, thanks to the gumbel-softmax trick \cite{jang2016gumbelsoftmax}, AutoGCL is end-to-end differentiable yet providing sufficient variances for contrastive samples generation. We further propose a joint training strategy to train the learnable view generators, the graph encoders, and the classifier in an end-to-end manner. The strategy includes the view similarity loss, the contrastive loss, and the classification loss. It makes the proposed view generators generate augmented graphs that have similar semantic information but with different topological properties. In Table \ref{tab-aug-compare}, we summarize the properties of existing graph augmentation methods, where AutoGCL dominates in the comparisons.

We conduct extensive graph classification experiments using semi-supervised learning, unsupervised learning, and transfer learning tasks to evaluate the effectiveness of AutoGCL. The results show that AutoGCL improves the state-of-the-art graph contrastive learning performances on most of the datasets. In addition, we visualize the generated graphs on MNIST-Superpixel dataset \cite{monti2017mnistsuperpix} and reveal that AutoGCL could better preserve semantic structures of the input data than existing pre-defined view generators.


Our contributions can be summarized as follows.
\begin{itemize}
    \item We propose a graph contrastive learning framework with learnable graph view generators embedded into an auto augmentation strategy. To the best of our knowledge, this is the first work to build learnable generative node-wise augmentation policies for graph contrastive learning. 
    
    \item We propose a joint training strategy for training the graph view generators, the graph encoder, and the graph classifier under the context of graph contrastive learning in an end-to-end manner.
    
    \item We extensively evaluate the proposed method on a variety of graph classification datasets with semi-supervised, unsupervised, and transfer learning settings. The t-SNE and view visualization results also demonstrate the effectiveness of our method.
\end{itemize}

\section{Related Work}

\subsection{Graph Neural Networks}
Denote a graph as $g=(V, E)$ where the node features are $\boldsymbol{x}_v$ for $v \in V$. In this paper, we focus on the graph classification task using Graph Neural Networks (GNNs). GNNs generate node-level embedding $\boldsymbol{h}_v$ through aggregating the node features $\boldsymbol{x}_v$ of its neighbors. Each layer of GNNs serves as an iteration of aggregation, such that the node embedding after the $k$-th layers aggregates the information within its $k$-hop neighborhood. The $k$-th layer of GNNs can be formulated as
\begin{small}
    \begin{align}
    \boldsymbol{a}_v^{(k)} &= \text{AGGREGATE}^{(k)}  ( \{ \boldsymbol{h}_u^{(k-1)} : u \in \mathcal{N}(v) \} ) \\
    \boldsymbol{h}_v^{(k)} &= \text{COMBINE}^{(k)} ( \boldsymbol{h}_v^{(k-1)}, \boldsymbol{a}_v^{(k)} )
    \end{align}
\end{small}
For the downstream tasks such as graph classification, the graph-level representation $\boldsymbol{z}_g$ is obtained via the READOUT function and MLP layers as
\begin{small}
    \begin{align}
    F(g) &= \text{READOUT}(\{ \boldsymbol{h}_n^{(k)}: v_n \in \mathcal{V} \} ) \\
    \boldsymbol{z}_g &= \text{MLP} (  F(g) )
    \end{align}
\end{small}
In this work we follow the existing graph contrastive learning literature to employ two state-of-the-art GNNs, \textit{i.e.}, GIN \cite{xu2018gin} and ResGCN \cite{chen2019gfn}, as our backbone GNNs. 

\subsection{Pre-training Graph Neural Networks}

Pre-training GNNs on graph datasets still remains a challenging task, since the semantics of graphs are not straightforward, and the annotation of graphs (proteins, chemicals, etc.) usually requires professional domain knowledge. It is very costly to collect large-scale and fine-annotated graph datasets like ImageNet \cite{krizhevsky2012imagenet}. An alternative way is to pre-train the GNNs in an unsupervised manner. The GAE \cite{kipf2016vgae} first explored the unsupervised GNN pre-training by reconstructing the graph topological structure. GraphSAGE \cite{hamilton2017graphsage} proposed an inductive way of unsupervised node embedding by learning the neighborhood aggregation function. Pretrain-GNN \cite{hu2019pretraingnn} conducted the first systematic large-scale investigation of strategies for pre-training GNNs under the transfer learning setting. It proposed self-supervised pre-training strategies to learn both the local and global features of graphs. However, the benefits of graph transfer learning may be limited and lead to negative transfer \cite{rosenstein2005negtransfer}, as the graphs from different domains actually differ a lot in their structures, scales and node/edge attributes. Therefore, many of the following works started to explore an alternative approach, \textit{i.e.}, the contrastive learning, for GNNs pre-training.

\subsection{Contrastive Learning}

In recent years, contrastive learning (CL) has received considerable attention among the self-supervised learning approaches, and a series of CL methods including SimCLR \cite{chen2020simclr} and MoCo-v2 \cite{chen2020mocov2} even outperform the supervised baselines. Through minimizing the contrastive loss \cite{hadsell2006clloss}, the views generated from the same input (\textit{i.e.}, positive view pairs) are pulled close in the representation space, while the views of different inputs (\textit{i.e.}, negative view pairs) are pushed apart. Most of the existing CL methods \cite{he2020moco,zbontar2021barlowtwins,chen2020simclr, grill2020byol} generate views using data augmentation, which is still challenging and under-explored for the graph data. Instead of data augmentation, DGI \cite{velivckovic2018dgi} treated the graph-level and node-level representations of the same graph as positive view pairs. CMRLG \cite{hassani2020cmcgnn} achieved an analogical goal by treating the adjacency matrix and the diffusion matrix as positive pairs. More recently, the GraphCL framework \cite{you2020graphcl} employed four types of graph augmentations, including node dropping, edge perturbation\footref{foot-edge-perturb}, sub-graph sampling\footref{foot-subgraph}, and node attribute masking\footref{foot-attr-mask}, enabling the most diverse augmentations by far for graph view generation. GCA \cite{zhu2020gca} used sub-graph sampling and node attribute masking as augmentations and introduced a prior augmentation probability based on the node centrality measures, enabling more adaptiveness than GraphCL \cite{you2020graphcl}, but the prior is not learnable.


\subsection{Learnable Data Augmentation}

As mentioned above, data augmentation is a significant component of CL. The existing literature \cite{chen2020simclr,you2020graphcl} has revealed that the optimal augmentation policies are task-dependent and the choice of augmentations makes a considerable difference to the CL performance. The researchers have explored to automatically discover the optimal policy for image augmentations in the computer vision field. For instance, AutoAugment \cite{cubuk2019autoaugment} firstly optimized the combination of augmentation functions through reinforcement learning. Faster-AA \cite{hataya2020fasteraa} and DADA \cite{li2020dada} proposed a differentiable augmentation optimization framework following the DARTS \cite{liu2018darts} style. 



However, the learnable data augmentation methods are barely explored for CL except the InfoMin framework \cite{tian2020goodview}, which claims that good views of CL should maintain the label information as well as minimizing the mutual information of positive view pairs. InfoMin employs a flow-based generative model as the view generator for data augmentation and trains the view generator in a semi-supervised manner. However, transferring this idea to graph is a non-trivial task since current graph generative models are either of limited generation qualities \cite{kipf2016vgae} or designed for specific tasks such as the molecular data \cite{de2018molgan, madhawa2019graphnvp, wang2021molecular}. To make graph augmentations adaptive to different tasks, JOAO \cite{you2021joao} learns the sampling distribution of pre-defined augmentations in a Bayesian manner, but the augmentations themselves are still not learnable. AD-GCL \cite{suresh2021adgcl} first proposed a learnable edge dropping augmentation and employs adversarial training strategy, but node-level augmentations are not considered, and the strategy will not ensure to generate label-preserving augmentations.

In this work we build a learnable graph view generator that learns a probability distribution over the node-level augmentations. Compared to the existing graph CL methods, our method well preserves the semantic structures of original graphs. Moreover, it is end-to-end differentiable and can be efficiently trained. 


\section{Methodology}

\subsection{What Makes a Good Graph View Generator?}
\label{sec-insight}

Our goal is to design a learnable graph view generator that learns to generate the augmented graph view in data-driven manner. Although various graph data augmentation methods have been proposed, there is less discussion on what makes a good graph view generator? From our perspective, an ideal graph view generator for data augmentation and contrastive learning should satisfy the following properties: (1) It supports both the augmentations of the graph \textbf{topology} and the \textbf{node feature}. (2) It is \textbf{label-preserving}, \textit{i.e.}, the augmented graph should maintain the semantic information in the original graph.
(3) It is \textbf{adaptive} to different data distributions and scalable to large graphs. (4) It provides \textbf{sufficient variances} for contrastive multi-view pre-training. (5) It is \textbf{end-to-end differentiable} and \textbf{efficient} enough for fast gradient computation via \textbf{back-propagation (BP)}.


Here we provide an overview of the augmentation methods proposed in existing literature of graph contrastive learning in Table \ref{tab-aug-compare}. CMRLG \cite{hassani2020cmcgnn} applies diffusion kernel to get different topological structures. GRACE \cite{zhu2020grace} uses random edge dropping and node attribute masking\footnote{Randomly mask the attributes of certain ratio of nodes. \label{foot-attr-mask}}. GCA \cite{zhu2020gca} uses node dropping and node attribute masking along with a structural prior. GraphCL \cite{you2020graphcl} proposes the most flexible set of graph data augmentations so far, including node dropping, edge perturbation\footnote{Randomly replace certain ratio of edges with random edges. \label{foot-edge-perturb}}, sub-graph\footnote{Randomly select a connected subgraph of certain size. \label{foot-subgraph}}, and attribute masking\footref{foot-attr-mask}. We provide a detailed ablation study and analysis of GraphCL augmentations with different augmentation ratios in Section 1.1 of the supplementary. JOAO \cite{you2021joao} optimizes the augmentation sampling policy of GraphCL in a Bayesian manner. AD-GCL \cite{suresh2021adgcl} designs a learnable edge dropping augmentation.

In this work, we propose a learnable view generator to address all the above issues. Our view generator includes both augmentations of node dropping and attribute masking, but it is much more flexible since both two augmentations can be simultaneously employed in a node-wise manner, without the need of tuning the \emph{``aug ratio''}. Besides the concern of model performance, another reason for not incorporating edge perturbation in our view generator is, the generation of edges through the learnable methods (\textit{e.g.}, VGAE \cite{kipf2016vgae}) requires to predict the full adjacency matrix that contains $O(N^2)$ elements, which is a heavy burden for back-propagation when dealing with large-scale graphs.

\subsection{Learnable Graph View Generator}

\begin{figure}[t]
    \begin{center}
    \includegraphics[width=1\linewidth]{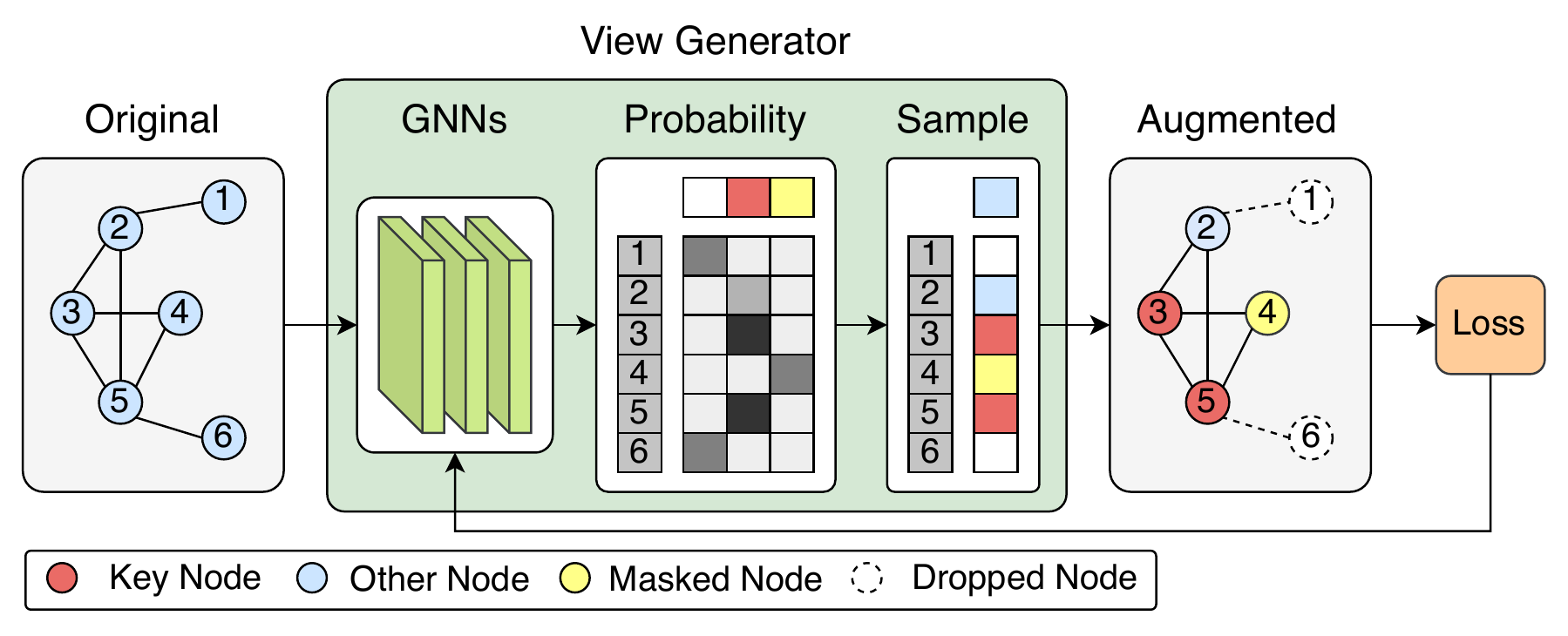}
    \end{center}
    \vspace{-0.5cm}
    \caption{The architecture of our learnable graph view generator. The GNN layers embed the original graph to generate a distribution for each node. The augmentation choice of each node is sampled from it using the gumbel-softmax.
    }
    \vspace{-0.5cm}
    \label{fig-view-generator}
\end{figure}

Figure \ref{fig-view-generator} illustrates the scheme of our proposed learnable graph view generator. We use GIN \cite{xu2018gin} layers to get the node embedding from the node attribute. For each node, we use the embedded node feature to predict the probability of selecting a certain augment operation. The augmentation pool for each node is drop, keep, and mask. We employ the gumbel-softamx \cite{jang2016gumbelsoftmax} to sample from these probabilities then assign an augmentation operation to each node. Formally, if we use $k$ GIN layers as the embedding layer, we denote $\boldsymbol{h}_v^{(k)}$ as the hidden state of node $v$ at the $k$-th layer and $\boldsymbol{a}_v^{(k)}$ as the embedding of node $v$ after the $k$-th layer. For node $v$, we have the node feature $\boldsymbol{x}_v $, the augmentation choice $f_v$, and the function $\text{Aug}(\boldsymbol{x}, f)$ for applying the augmentation. Then the augmented feature $\boldsymbol{x}_v^{'}$ of node $v$ is obtained via
\begin{small}
    \begin{align}
        \boldsymbol{h}_v^{(k-1)} &= \text{COMBINE}^{(k)} ( \boldsymbol{h}_v^{(k-2)}, \boldsymbol{a}_v^{(k-1)} ) \\
        \boldsymbol{a}_v^{(k)} &= \text{AGGREGATE}^{(k)}  ( \{ \boldsymbol{h}_u^{(k-1)} : u \in \mathcal{N}(v) \} ) \\
        f_v &= \text{GumbelSoftmax} ( \boldsymbol{a}_v^{(k)} ) \\
        \boldsymbol{x}_v^{'} &=  \text{Aug}(\boldsymbol{x}_v, f_v)
    \end{align}
\end{small}
The dimension of the last layer k is set as the same number of possible augmentations for each node. $\boldsymbol{a}_v^{(k)}$ denotes the probability distribution for selecting each kind of augmentation. $f_v$ is a one-hot vector sampled from this distribution via gumbel-softmax and it is differentiable due to the reparameterization trick. The augmentation applying function $\text{Aug}(\boldsymbol{x}_v, f_v)$ combines the node attribute $\boldsymbol{x}_v$ and $f_v$ using differentiable operations (e.g. multiplication), so the gradients of the weights of the view generator are kept in the augmented node features and can be computed using back-propagation. For the augmented graph, the edge table is updated using $f_v$ for all $v\in V$, where the edges connected to any dropped nodes are removed. As the edge table is only the guidance for node feature aggregation and it does not participate in the gradient computation, it does not need to be updated in a differentiable manner. Therefore, our view generator is end-to-end differentiable. The GIN embedding layers and the gumbel-softmax can be efficiently scaled up for larger graph datasets and more augmentation choices.

\subsection{Contrastive Pre-training Strategy}

Since the contrastive learning requires multiple views to form a positive view pair, we have two view generators and one classifier for our framework. According to InfoMin principle \cite{tian2020goodview}, a good positive view pair for contrastive learning should maximize the label-related information as well as minimizing the mutual information (similarity) between them. To achieve that, our framework uses two separate graph view generators and trains them and the classifier in a joint manner.

\begin{figure*}[t]
    \begin{center}
    \includegraphics[width=1\textwidth]{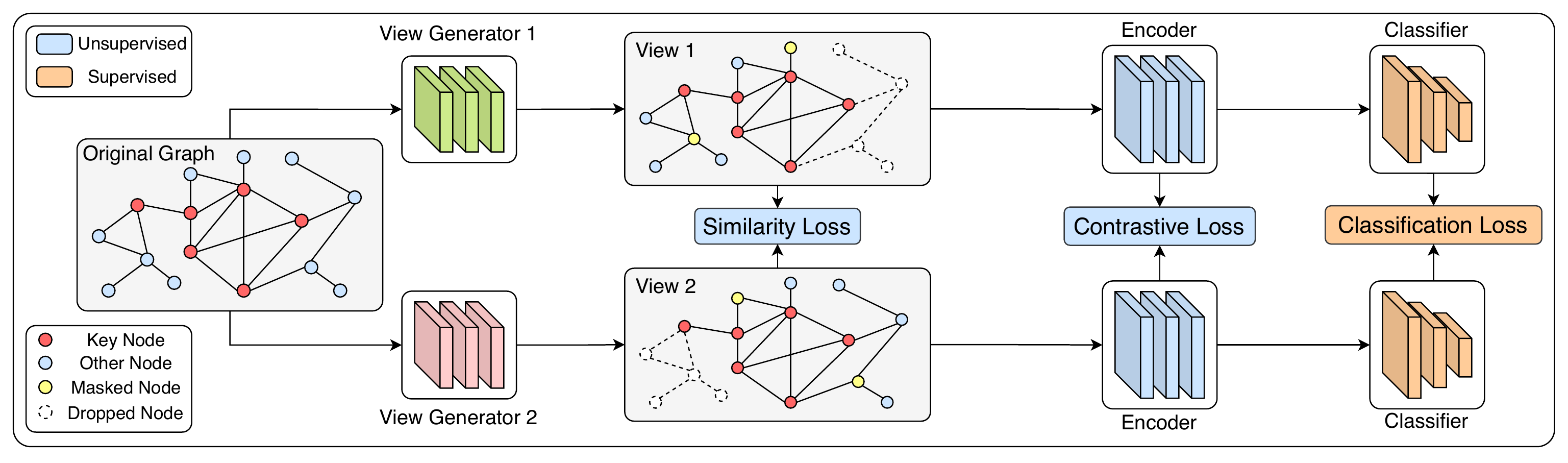}
    \end{center}
    \vspace{-0.6cm}
    \caption{The proposed AutoGCL framework is composed of three parts: (1) two \emph{view generators} that generate different views of the original graph, (2) a \emph{graph encoder} that extracts the features of graphs and (3) a \emph{classifier} that provides the graph outputs.}
    \vspace{-0.6cm}
    \label{fig-framework}
\end{figure*}

\subsubsection{Loss Function Definition}

Here we define three loss functions, contrastive loss $\mathcal{L}_{\text{cl}}$, similarity loss $\mathcal{L}_{\text{sim}}$, and classification loss $\mathcal{L}_{\text{cls}}$. For contrastive loss, we follow the previous works \cite{chen2020simclr,you2020graphcl} and use the normalized temperature-scaled cross entropy loss (NT-XEnt) \cite{sohn2016clloss}. Define the similarity function $\text{sim}(\boldsymbol{z}_1,\boldsymbol{z}_2)$ as
\begin{small}
    \begin{align}
    \text{sim}(\boldsymbol{z}_1,\boldsymbol{z}_2) = \frac{\boldsymbol{z}_1 \cdot \boldsymbol{z}_2}{ {\lVert \boldsymbol{z}_1 \rVert}_2 \cdot {\lVert \boldsymbol{z}_2 \rVert}_2 }
    \end{align}
\end{small}
Suppose we have a data batch made up of $N$ graphs. We pass the batch to the two view generators to obtain $2N$ graph views. We regard the two augmented views from the same input graph as the positive view pair. We use $\mathbbm{1}_{[k \neq i]} \in \{0, 1\}$ to denote the indicator function. We denote the contrastive loss function for a positive pair of samples $(i, j)$ as $\ell(i,j)$, the contrastive loss of this data batch as $\mathcal{L}_{\text{cl}}$, the temperature parameter as $\tau$, then we have
\begin{small}
    \begin{align}
    \ell_{(i,j)} &= - \log \frac{\exp ( \text{sim} (\boldsymbol{z}_i, \boldsymbol{z}_j) / \tau )}{ \sum_{k=1}^{2N} \mathbbm{1}_{[k \neq i]} \exp(\text{sim} (\boldsymbol{z}_i, \boldsymbol{z}_k) / \tau ) } \\
    \mathcal{L}_{\text{cl}} &= \frac{1}{2N} \sum_{k=1}^{N} [\ell(2k-1, 2k) + \ell(2k, 2k-1)]
    \end{align}
\end{small}
The similarity loss is used to minimize the mutual information between the views generated by the two view generators. During the view generation process, we have a sampled state matrix $S$ indicting each node's corresponding augmentation operation (see Figure \ref{fig-view-generator}). For a graph $G$, we denote the sampled augmentation choice matrix of each view generator as $A_1, A_2$, then we formulate the similarity loss $\mathcal{L}_{\text{sim}}$ as
\begin{small}
    \begin{align}
    \mathcal{L}_{\text{sim}} = \text{sim}(A_1, A_2)
    \end{align}
\end{small}
Finally, for the classification loss, we directly use the cross entropy loss ($\ell_\text{cls}$). For a graph sample $g$ with class label $y$, we denote the augmented view as $g_1$ and $g_2$ and the classifier as $F$. Then the classification loss $\mathcal{L}_{\text{cls}}$ is formulated as
\begin{small}
    \begin{align}
    \mathcal{L}_{\text{cls}} &= \ell_\text{cls}(F(g), y) + \ell_\text{cls}(F(g_1), y) + \ell_\text{cls}(F(g_2), y)
    \end{align}
\end{small}
$\mathcal{L}_{\text{cls}}$ is employed in the semi-supervised pre-training task to encourage the view generator to generate label-preserving augmentations. 

\subsubsection{Naive Training Strategy}

For unsupervised learning and transfer learning tasks, we use a naive training strategy (naive-strategy). Since we do not know the label in the pre-training stage, the $\mathcal{L}_{\text{sim}}$ is not used because it does not make sense to just encourage the views to be different without keeping the label-related information. This could lead to generating useless or even harmful view samples. We just train the view generators and the classifier jointly to minimize the $\mathcal{L}_{\text{cl}}$ in the pre-training stage. 

Also, we note that the quality of the generated views will not be as good as the original data. During the $\mathcal{L}_{\text{cl}}$ minimization, instead of just minimizing the $\mathcal{L}_{\text{cl}}$ between two augmented views like GraphCL \cite{you2020graphcl}, we also make use of the original data. By pulling the original data and the augmented views close in the embedding space, the view generator are encouraged to preserve the label-related information. The details are described in Algorithm \ref{algo-naive}.

\setlength{\textfloatsep}{10pt} 
\begin{algorithm}[tb]
\begin{small}
    \caption{Naive training strategy (naive-strategy).}
        \begin{algorithmic}[1]
            \State Initialize weights of the two view generator $G_1$, $G_2$
            \State Initialize weights of the classifer $F$
            \While{not reached maximum epochs}
                \For{mini-batch $x$ from unlabeled data}
                    \State Get augmentation $x_1 = G_1(x), x_2 = G_2(x)$
                    \State Sample two views $v_1, v_2$ from $\{x, x_1, x_2\}$
                    \State $\mathcal{L} = \mathcal{L}_{\text{cl}} (v_1, v_2)$
                    \State Update the weights of $G_1, G_2, F$ to minimize $\mathcal{L}$
                \EndFor
            \EndWhile
            \While{not reached maximum epochs}
                \For{mini-batch $x$ from labeled data}
                    \State $\mathcal{L} = \mathcal{L}_{\text{cls}}(x)$
                    \State Update the weights of $F$ to minimize $\mathcal{L}$
                \EndFor
            \EndWhile
        \end{algorithmic}
    \label{algo-naive}
\end{small}
\vspace{-0.1cm}
\end{algorithm}

\subsubsection{Joint Training Strategy}

For semi-supervised learning tasks, we proposed a joint training strategy, performs contrastive training and supervised training alternately. This strategy generates label-preserving augmentation and outperforms the naive-strategy, the experiment results and detailed analysis is shown in Section \ref{sec-semi-exp} and Section \ref{sec-joint-strategy-analysis}.

For the joint-strategy, during the unsupervised training stage, we fix the view generators, and train the classifer by contrastive learning using unlabeled data. During the supervised training stage, we jointly train the view generator with the classifier using labeled data. By simultaneously optimizing $\mathcal{L}_{\text{sim}}$ and $\mathcal{L}_{\text{cls}}$, the two view generator are encouraged to generated label-preserving augmentations, yet being different enough from each other. The unsupervised training stage and supervised training stage are repeated alternately. This is very different from previous graph contrastive learning methods. Previous work like GraphCL \cite{you2020graphcl} use the pre-training/fine-tuning strategy, which first minimizes the contrastive loss ($\mathcal{L}_{\text{cl}}$) until convergence using the unlabeled data and then fine-tunes it with the labeled data. 

\begin{algorithm}[tb]
    \begin{small}
        \caption{Joint training strategy (joint-strategy).}
        \begin{algorithmic}[1]
            \State Initialize weights of $G_1$, $G_2$, $F$.
            \While{not reached maximum epochs}
                \For{mini-batch $x$ from unlabeled data}
                    \State Fix the weights of $G_1, G_2$
                    \State Get augmentation $x_1 = G_1(x), x_2 = G_2(x)$
                    \State Sample two views $v_1, v_2$ from $\{x, x_1, x_2\}$
                    \State $\mathcal{L} = \mathcal{L}_{\text{cl}} (v_1, v_2)$
                    \State Update the weights of $F$ to minimize $\mathcal{L}$
                \EndFor
                \For{mini-batch $x$ from labeled data}
                    \State Get augmentation $x_1 = G_1(x), x_2 = G_2(x)$
                    \State $\mathcal{L} = \mathcal{L}_{\text{cls}}(x, x_1, x_2) + \lambda \cdot \mathcal{L}_{\text{sim}}(x_1, x_2)$
                    \State Update the weights of $G_1, G_2, F$ to minimize $\mathcal{L}$
                \EndFor
            \EndWhile
        \end{algorithmic}
        \label{algo-joint}
    \end{small}
\vspace{-0.1cm}
\end{algorithm}
However, we found that for graph contrastive learning, the pre-training/fine-tuning strategy are more likely to cause over-fitting in the fine-tuning stage. And minimizing the $\mathcal{L}_{\text{cl}}$ too much may have negative effect for the fine-tuning stage (see Section \ref{sec-joint-strategy-analysis}). We speculate that minimizing the $\mathcal{L}_{\text{cl}}$ too much will push data points near the decision boundary to be too closed to each other, thus become more difficult the classifer to separate them. Because no matter how well we train the GNN classifer, there are still mis-classified samples due to the natural overlaps between the data distribution of different classes. But in the contrastive pre-training state, the classifer is not aware of whether the samples being pulled together are really from the same class.

\begin{table*}
        \caption{Comparison with the existing methods for unsupervised learning. The \textbf{bold} numbers denote the best performance and the the numbers in \textcolor{blue}{blue} represent the second best performance.}
        \vspace{-0.3cm}
        \fontsize{5}{5.5}\selectfont
        \resizebox{1\textwidth}{!}{
            \begin{tabular}{ccccccccc}
            \hline
            Model   & MUTAG       & PROTEINS   & DD & NCI1       & COLLAB & IMDB-B     & REDDIT-B   & REDDIT-M-5K \\
            \hline
            GL        & 81.66±2.11  & -          & -  & -          & -      & 65.87±0.98 & 77.34±0.18 & 41.01±0.17  \\
            WL        & 80.72±3.00  & 72.92±0.56 & -  & 80.01±0.50 & -      & 72.30±3.44 & 68.82±0.41 & 46.06±0.21  \\
            DGK       & 87.44±2.72  & 73.30±0.82 & -  & \textcolor{blue}{80.31±0.46} & -      & 66.96±0.56 & 78.04±0.39 & 41.27±0.18  \\
            \hline
            node2vec  & 72.63±10.20 & 57.49±3.57 & -  & 54.89±1.61 & -      & -          & -          & -           \\
            sub2vec   & 61.05±15.80 & 53.03±5.55 & -  & 52.84±1.47 & -      & 55.26±1.54 & 71.48±0.41 & 36.68±0.42  \\
            graph2vec & 83.15±9.25  & 73.30±2.05 & -  & 73.22±1.81 & -      & 71.10±0.54 & 75.78±1.03 & 47.86±0.26  \\
            \hline
            InfoGraph       & \textbf{89.01±1.13} & \textcolor{blue}{74.44±0.31} & 72.85±1.78 & 76.20±1.06 & 70.65±1.13 & \textcolor{blue}{73.03±0.87} & 82.50±1.42 & 53.46±1.03 \\
            GraphCL  & 86.80±1.34 & 74.39±0.45 & \textbf{78.62±0.40} & 77.87±0.41 & \textcolor{blue}{71.36±1.15} & 71.14±0.44 & \textbf{89.53±0.84} & \textcolor{blue}{55.99±0.28} \\
            JOAOv2  & - & 71.25±0.85 & 66.91±1.75 & 72.99±0.75 & 70.40±2.21 & 71.60±0.86 & 78.35±1.38 & 45.57±2.86 \\
            AD-GCL  & - & 73.59±0.65 & 74.49±0.52 & 69.67±0.51 & \textbf{73.32±0.61} & 71.57±1.01 & 85.52±0.79 & 53.00±0.82 \\
            Ours & \textcolor{blue}{88.64±1.08} & \textbf{75.80±0.36} & \textcolor{blue}{77.57±0.60} & \textbf{82.00±0.29} & 70.12±0.68 & \textbf{73.30±0.40} & \textcolor{blue}{88.58±1.49} & \textbf{56.75±0.18} \\
            \hline
            \end{tabular}
        }
    \vspace{-0.3cm}
    \label{tab-unsup-exp}
\end{table*}

\begin{table*}
    \caption{Comparison with the existing methods for transfer learning. The \textbf{bold} numbers denote the best performance and the numbers in \textcolor{blue}{blue} denote the second best performance.}
    \vspace{-0.3cm}
    \fontsize{5}{5.5}\selectfont
    \resizebox{1\textwidth}{!}{
        \begin{tabular}{lcccccccc}
        \hline
        Model                 & BBBP       & Tox21      & ToxCast    & SIDER      & ClinTox    & MUV        & HIV        & BACE       \\
        \hline
        \emph{No  Pretrain}     & 65.8±4.5   & 74.0±0.8   & 63.4±0.6   & 57.3±1.6   & 58.0±4.4   & 71.8±2.5   & 75.3±1.9   & 70.1±5.4   \\ \hline
        Infomax                 & 68.8±0.8   & 75.3±0.5   & 62.7±0.4   & 58.4±0.8   & 69.9±3.0   & 75.3±2.5   & 76.0±0.7   & 75.9±1.6   \\
        EdgePred                & 67.3±2.4   & 76.0±0.6   & \textcolor{blue}{64.1±0.6}   & 60.4±0.7   & 64.1±3.7   & 74.1±2.1   & 76.3±1.0   & \textcolor{blue}{79.9±0.9}   \\
        AttrMasking             & 64.3±2.8   & \textbf{76.7±0.4}   & \textbf{64.2±0.5}   & 61.0±0.7   & 71.8±4.1   & 74.7±1.4   & 77.2±1.1   & 79.3±1.6   \\
        ContextPred             & 68.0±2.0   & 75.7±0.7   & 63.9±0.6   & 60.9±0.6   & 65.9±3.8   & \textcolor{blue}{75.8±1.7}   & 77.3±1.0   & 79.6±1.2   \\
        GraphCL                 & 69.68±0.67 & 73.87±0.66 & 62.40±0.57 & 60.53±0.88 & 75.99±2.65 & 69.80±2.66 & \textbf{78.47±1.22} & 75.38±1.44 \\
        JOAOv2                 & \textcolor{blue}{71.39±0.92} & 74.27±0.62 & 63.16±0.45 & 60.49±0.74 & 80.97±1.64 & 73.67±1.00 & 77.51±1.17 & 75.49±1.27 \\
        AD-GCL                 & 70.01±1.07 & \textcolor{blue}{76.54±0.82} & 63.07±0.72 & \textbf{63.28±0.79} & \textcolor{blue}{79.78±3.52} & 72.30±1.61 & 78.28±0.97 & 78.51±0.80 \\
        Ours & \textbf{73.36±0.77} & 75.69±0.29 & 63.47±0.38 & \textcolor{blue}{62.51±0.63} & \textbf{80.99±3.38} & \textbf{75.83±1.30} & \textcolor{blue}{78.35±0.64} & \textbf{83.26±1.13} \\
        \hline
        \end{tabular}
    }
    \vspace{-0.6cm}
    \label{tab-transfer-exp}
\end{table*}

Therefore, we propose a new semi-supervised training strategy, namely the joint-strategy by alternately minimizing the $\mathcal{L}_{\text{cl}}$ and $\mathcal{L}_{\text{cls}} + \mathcal{L}_{\text{cls}}$. Minimizing $\mathcal{L}_{\text{cls}} + \mathcal{L}_{\text{cls}}$ is inspired by InfoMin \cite{tian2020goodview}, so as to make the two view generator to keep label-related information while having less mutual information. 
However, since we only have a small portion of labeled data to train our view generator, it is still beneficial to use the original data just like the naive-strategy. Interestingly, since we need to minimize $\mathcal{L}_{\text{cls}}$ and $\mathcal{L}_{\text{sim}}$ simultaneously, a weight $\lambda$ can be applied to better balance the optimization, but actually we found setting $\lambda=1$ works pretty well during the experiments in Section \ref{sec-sota}. The detailed training strategy is described in Algorithm \ref{algo-joint}. And the overview of our whole framework is shown in Figure \ref{fig-framework}.

\section{Experiment}

\subsection{Comparison with State-of-the-Art Methods}
\label{sec-sota}
    
\subsubsection{Unsupervised Learning}

For the unsupervised graph classification task, we contrastively train a representation model using unlabeled data, then fix the representation model and train the classifier using labeled data. Following GraphCL \cite{you2020graphcl}, we use a 5-layer GIN with a hidden size of 128 as our representation model, and use an SVM as our classifier. We train the GIN with a batch size of 128 and a learning rate of 0.001. There are 30 epochs of contrastive pre-training under the naive-strategy. We perform a 10-fold cross validation on every dataset. For each fold, we employ 90\% of the total data as the unlabeled data for contrastive pre-training, and 10\% as the labeled testing data. We repeat every experiment for 5 times using different random seeds. 

We compare with the kernel-based methods like graphlet kernel (GL) \shortcite{shervashidze2009graphlet}, Weisfeiler-Lehman sub-tree kernel (WL) \shortcite{shervashidze2011wl} and deep graph kernel (DGK) \shortcite{yanardag2015dgk}, and other unsupervised graph representation methods like node2vec \cite{grover2016node2vec}, sub2vec \cite{adhikari2018sub2vec}, graph2vec \cite{narayanan2017graph2vec} also the contrastive learning methods like InfoGraph \cite{sun2019infograph}, GraphCL \cite{you2020graphcl}, JOAO \cite{you2021joao} and AD-GCL \cite{suresh2021adgcl}. Table \ref{tab-unsup-exp} show the comparison among different models for unsupervised learning. Our proposed model achieves the best results on PROTEINS, NCI1, IMDB-binary, and REDDIT-Multi-5K datasets and the second best performances on MUTAG, DD, and REDDIT-binary datasets, outperforming current state-of-the-art contrastive learning methods GraphCL, JOAO and AD-GCL.

\subsubsection{Transfer Learning}

We also evaluate the transfer learning performance of the proposed method. A strong baseline method for graph transfer learning is Pretrain-GNN \cite{hu2019pretraingnn}. The network backbone of Pretrain-GNN, GraphCL, JOAO, AD-GCL and our method is a variant of GIN \cite{xu2018gin}, which incorporates the edge attribute. We perform 100 epochs of supervised pre-training on the pre-processed ChEMBL dataset (\cite{mayr2018chemdataset, gaulton2012chembl}), which contains 456K molecules with 1,310 kinds of diverse and extensive biochemical assays. 

We perform 30 epochs of fine-tuning on the 8 chemistry evaluation subsets. We use a hidden size of 300 for the classifier, a hidden size of 128 for the view generator. We train the model using a batch size of 256 and a learning rate of 0.001. The results in Table \ref{tab-transfer-exp} are the mean±std of the ROC-AUC scores from 10 reps. Infomax, EdgePred, AttrMasking, ContextPred are the manually designed pre-training strategies from Pretrain-GNN \cite{hu2019pretraingnn}. 

Table \ref{tab-transfer-exp} presents the comparison among different methods. Our proposed method achieves the best performance on most dataset, such as  BBBP, ClinTox, MUV, and BACE, and compared with the current SoTA model AD-GCL \cite{suresh2021adgcl}, our method performs considerably better, for example, on BACE dataset, the accuracy raises from 78.51±0.80 to 83.26±1.13. Considering all datasets, the average gain of using our proposed method is around 1.5\%. Interestingly, AttrMasking achieves the best performance on Tox21 and ToxCast, which is slightly better than our method. One possible reason is that attributes are important for classification in Tox21 and ToxCast datasets.

\subsubsection{Semi-Supervised Learning}
\label{sec-semi-exp}

\begin{table*}
    \caption{Comparison with existing methods and different strategies for semi-supervised learning. \textbf{Bold} numbers denote the best performance and the numbers in \textcolor{blue}{blue} denote the second best performance. \textcolor{red}{Red} is our default setting for joint training strategy.}
    \vspace{-0.3cm}
    \resizebox{1\textwidth}{!}{
        \begin{tabular}{lcccccccc}
        \hline
        Model                   & PROTEINS   & DD         & NCI1       & COLLAB     & GITHUB     & IMDB-B     & REDDIT-B   & REDDIT-M-5K \\
        \hline
        \emph{Full Data}        & 78.25±1.61 & 80.73±3.78 & 83.65±1.16 & 83.44±0.77 & 66.89±1.04 & 76.60±4.20 & 89.95±2.06 & 55.59±2.24  \\ 
        \hline
        10\% Data               & 69.72±6.71 & 74.36±5.86 & 75.16±2.07 & 74.34±2.00 & 61.05±1.57 & 64.80±4.92 & 76.75±5.60 & 49.71±3.20  \\
        10\%   GCA              & 73.85±5.56 & 76.74±4.09 & 68.73±2.36 & 74.32±2.30 & 59.24±3.21 & \textbf{73.70±4.88} & 77.15±6.96 & 32.95±10.89 \\
        10\%   GraphCL Aug Only & 70.71±5.63 & 76.48±4.12 & 70.97±2.08 & 73.56±2.52 & 59.80±1.94 & 71.10±5.11 & 76.45±4.83 & 47.33±4.02 \\
        10\%   GraphCL CL       & 74.21±4.50 & 76.65±5.12 & 73.16±2.90 & 75.50±2.15 & \textcolor{blue}{63.51±1.02} & 68.10±5.15 & 78.05±2.65 & 48.09±1.74 \\
        10\%   JOAOv2           & 73.31±0.48 & 75.81±0.73 & 74.86±0.39 & 75.53±0.18 & \textbf{66.66±0.60} & - & \textcolor{blue}{88.79±0.65} & \textcolor{blue}{52.71±0.28} \\
        10\%   AD-GCL           & 73.96±0.47 & \textbf{77.91±0.73} & \textbf{75.18±0.31} & 75.82±0.26 & - & - & \textbf{90.10±0.15} & \textbf{53.49±0.28} \\
        10\%   Our Aug Only     & \textcolor{blue}{75.49±5.15} & 77.16±4.53 & 73.33±2.86 & 75.92±1.93 & 60.65±1.04 & \textcolor{blue}{71.90±2.88} & 79.65±2.84 & 47.97±2.22 \\
        10\%   Our CL Naive     & 74.57±3.29 & 75.55±4.76 & 73.22±2.48 & \textcolor{blue}{76.60±2.15} & 60.95±1.32 & 71.00±2.91 & 79.10±4.38 & 46.71±2.64 \\
        10\%   Our CL Joint ($\mathcal{L}_\text{cls}$) & 74.66±2.58 & 76.57±5.08 & 71.78±1.61 & 75.38±2.15 & 60.39±1.50 & 70.60±4.17 & 78.90±3.11 & 46.89±3.13 \\
        10\%   Our CL Joint ($\mathcal{L}_\text{cls}$+$\mathcal{L}_\text{sim}$)     & 75.12±3.35 & 76.23±3.57 & 72.55±2.72 & 75.60±2.08 & 60.18±1.75 & 71.70±3.86 & 79.25±2.88 & 47.51±2.51 \\
        10\%   Our CL Joint ($\mathcal{L}_\text{cl}$ +$\mathcal{L}_\text{cls}$ )     & 74.75±3.35 & 76.82±3.85 & 73.07±2.31 & 76.18±2.46 & 61.75±1.30 & 71.50±5.32 & 78.35±4.21 & 47.73±2.69 \\
        10\%   \textcolor{red}{Our CL Joint} ($\mathcal{L}_\text{cl}$ +$\mathcal{L}_\text{cls}$+$\mathcal{L}_\text{sim}$)     & \textbf{75.65±2.40} & \textcolor{blue}{77.50±4.41} & \textcolor{blue}{73.75±2.25} & \textbf{77.16±1.48} & 62.46±1.51 & \textcolor{blue}{71.90±4.79} & 79.80±3.47 & 49.91±2.70 \\
        \hline
        \end{tabular}
        }
    \vspace{-0.5cm}
    \label{tab-semi-exp}
\end{table*}

We perform semi-supervised graph classification task on TUDataset \cite{morris2020tudataset}. 
For our view generator, we use a 5-layer GIN with a hidden size of 128 as the embedding model. We use ResGCN \shortcite{chen2019gfn} with a hidden size of 128 as the classifier. For GraphCL, we use the default augmentation policy \emph{random4}, which randomly selects two augmentations from node dropout, edge perturbation, subgraph, and attribute masking for every mini-batch. All augmentation ratios are set to $0.2$, which is also the default setting in GraphCL.


We employ a 10-fold cross validation on each dataset. For each fold, we use 80\% of the total data as the unlabeled data, 10\% as labeled training data, and 10\% as labeled testing data. For the augmentation only (Aug Only) experiments, we only perform 30 epochs of supervised training with augmentations using labeled data. For the contrastive learning experiments of GraphCL and our naive-strategy, we perform 30 epochs of contrastive pre-training followed by 30 epochs of supervised training. For our joint-strategy, there is 30 joint epochs of contrastive training and supervised training. 

Table \ref{tab-semi-exp} compares the performances obtained by different training strategies: augmentation only (Aug only), naive-strategy 
(CL naive) and joint-strategy (CL joint). We also conducted an ablation study of our joint loss function. The proposed CL joint approach achieves relatively high accuracy on most datasets, for example, on PROTEINS and COLLAB datasets, using joint strategy obtains the best performance. In terms of other datasets, using joint strategy could also achieves the second best performances. Looking at the comparison among Aug only, CL naive and CL joint, CL joint is superior to the other two approaches, in particular to CL naive.

\subsection{Effectiveness of Learnable View Generators}
\label{sec-mnist-exp}

In this section, we demonstrate the superiority of learnable graph augmentation policies over the fixed ones. Since the graph datasets are usually difficult to be manually classified and visualized, we trained a view generator on MNIST-Superpixel dataset \cite{monti2017mnistsuperpix} to verify that our graph view generator is able to effectively capture the semantic information in graphs than GraphCL \cite{you2020graphcl}, since MNIST-Superpixel graphs have clear semantics which does not require any domin knowledge. The visualization result is shown in Figure \ref{fig-mnist-superpixel-vis}.



Here we jointly trained the view generators with the classifier until the test accuracy (evaluated on generated views) reached $90\%$. Since our only topological augmentation is node dropping. So we compared the view of GraphCL's node dropping augmentation, and use the default setting $\text{aug\_ratio}=2$. Figure \ref{fig-mnist-superpixel-vis} shows that, our view generator are more likely to keep key nodes in the original graph, preserving its semantic feature, yet providing enough variance for contrastive learning. Details of the MNIST-Superpixel dataset and more visualization examples are shown in Section 1.2 of the supplementary.

\begin{figure}[tbp]
    \begin{center}
    \includegraphics[width=1\linewidth]{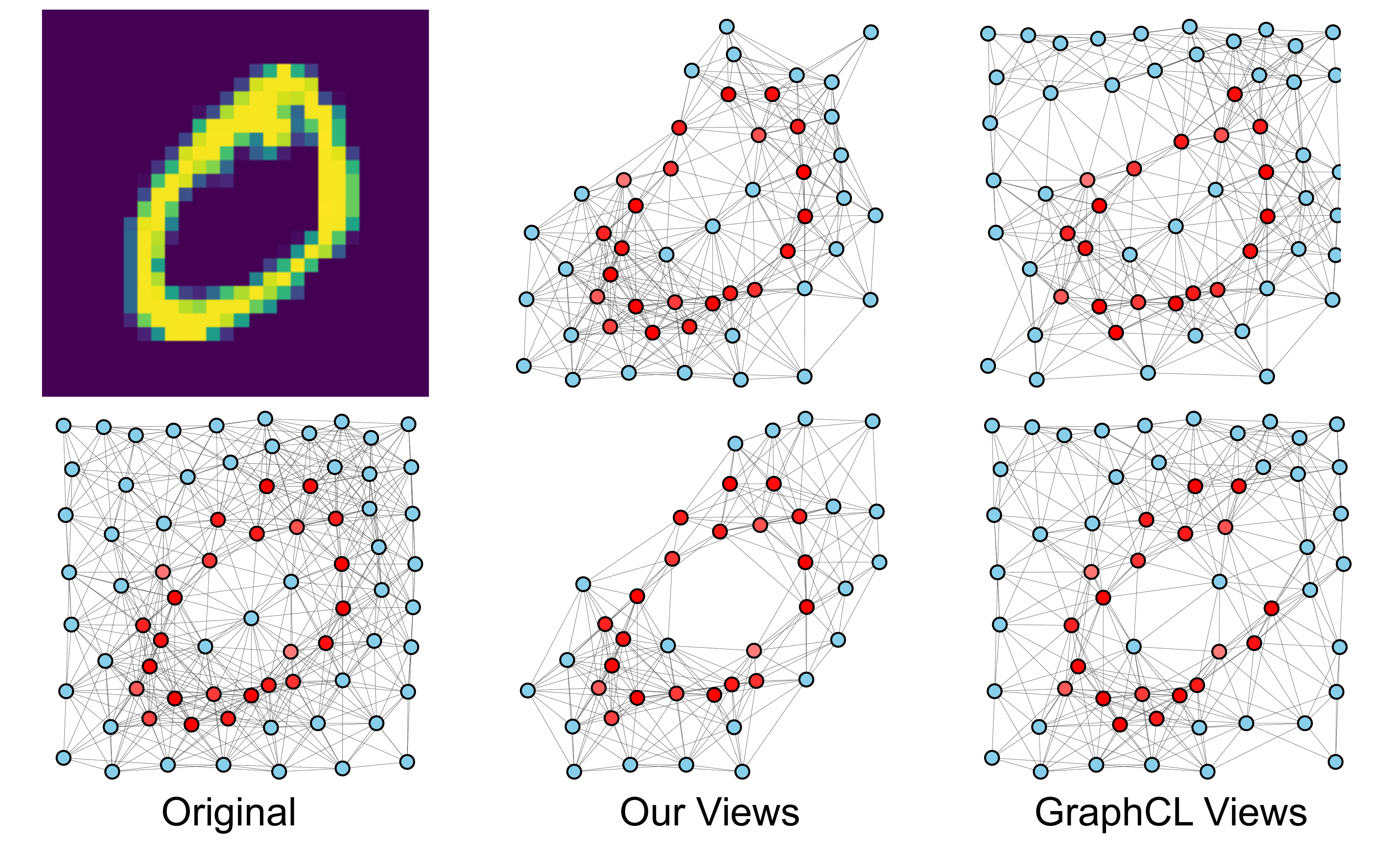}
    \end{center}
    \vspace{-0.4cm}
    \caption{View visualization on the MNIST-Superpixel dataset. Redness reflects the magnitude of node attribute.}
    \vspace{-0.1cm}
    \label{fig-mnist-superpixel-vis}
\end{figure}

\subsection{Analysis for Joint Training Strategy}
\label{sec-joint-strategy-analysis}

We compared the naive-strategy (Algorithm \ref{algo-naive}) with the joint-strategy (Algorithm \ref{algo-joint}). We trained on COLLAB \shortcite{collab} dataset, which have 5000 social network graphs of 3 classes, the average nodes and edges are 74.49 and 2457.78. Here we use 5-layer GIN \cite{xu2018gin} as the backbone for both the view generator and the classifier. For naive-strategy, there is 30 epochs of contrastive pretrain using 80\% unlabeled data and 30\% of fine-tuning using 10\% of data. For joint-strategy, there is 30 epochs of joint training. The learning curves are shown in Section 1.3 of the supplementary. Our results show that the joint strategy considerably alleviate the over-fitting effect, and our label-preserving view generator is very effective. We also visualize the process for learning the embedding for each strategy using t-SNE \cite{van2008tsne} in the supplementary. We found that joint-strategy leads to better representation much faster since labeled data is used for supervision, also this supervision signal could benefit view generator learning.

\section{Conclusion}
In this paper, we presented a learnable data augmentation approach to graph contrastive learning, where we employed GIN to generate different views of the original graphs. To preserve the semantic label of the input graph, we developed a joint learning strategy, which alternately optimizes view generators, graph encoders and the classifier. We also conducted extensive experiments on a number of datasets and tasks, such as semi-supervised learning, unsupervised learning and transfer learning, where results demonstrate the advantage of our proposed method outperforming counterparts on most datasets and tasks. In addition, we visualized the generated graph views, which could preserve discriminative structures of input graphs, benefiting classification. Finally, the t-SNE visualization illustrated that the proposed joint training strategy could be a better choice for semi-supervised graph representation learning.

\appendix

\section{More Analysis}

\subsection{An Insight into GraphCL Augmentations}

Here we want to prove that the augmentation selection policy and the intensity of augmentations really matter to the final results. Among all the previous works, GraphCL \cite{you2020graphcl} enables the most flexible set of graph data augmentations so far, as it includes node dropping, edge perturbation, sub-graph, and attribute masking. Where

\begin{itemize}
    \item Node dropping randomly removes certain ratio of nodes.
    \item Edge perturbation first randomly removes certain ratio of edges, then randomly adds the same number of edges. 
    \item Sub-graph randomly choosing a connected subgraph by firstly choose a random center node, then gradually add its neighbor nodes until certain ratio of the total nodes are reached. 
    \item  Node attribute masking randomly masks the attributes of certain ratio of nodes.
\end{itemize}

We note that the only augmentation selection policy of all existing works is uniform sampling and all the augmentation methods require a hyper-parameter \emph{``aug ratio''} that controls the portion of nodes/edges that are selected for augmentation. The \emph{``aug ratio''} is set to a constant in every experiment (\textit{e.g.}, 20\% by GraphCL's default). We perform an ablation study of these augmentation methods as shown in Table \ref{tab-graphcl-aug-select-ablation}, Table \ref{tab-graphcl-aug-ratio-ablation} and conclude that:

\begin{itemize}
    \item The positive contributions of edge perturbation and subgraph augmentation for graph contrastive learning are very limited (or even negative).
    \item The subgraph augmentation is actually contained in the augmentation space of node dropping. For instance, the potential view space of dropping 80\% of nodes contains the potential view space of selecting a connected subgraph that contains 20\% of the nodes. 
    \item The choice of \emph{``aug ratio''} has a considerable effect on the final performance. It is inappropriate to apply the same \emph{``aug ratio''} to different augmentations, datasets, and tasks. 
\end{itemize}
    
\begin{table}[h]
    \caption{Ablation Study of GraphCL Augmentations.}
    \begin{minipage}{1\linewidth}
    \centering
    \resizebox{1\hsize}{!}{
        \begin{tabular}{lcccc}
            \hline
            Augmentation            & NCI1         & PROTEINS     & DD           \\
            \hline
            \emph{Full   Data}       & 83.28 ± 1.84 & 77.18 ± 2.71 & 79.80 ± 2.65 \\
            10\%   Data        & 72.99 ± 1.28 & 67.50 ± 7.81 & 72.91 ± 4.42 \\
            10\%   Random4     & 72.94 ± 3.23 & 71.08 ± 5.03 & 75.29 ± 2.74 \\
            10\%   NodeDrop    & 72.55 ± 2.43 & 71.98 ± 4.29 & 76.66 ± 2.33 \\
            10\%   EdgePerturb & 71.85 ± 3.45 & 70.72 ± 5.60 & 73.34 ± 3.22 \\
            10\%   Subgraph    & 72.70 ± 2.60 & 62.99 ± 4.92 & 68.33 ± 4.77 \\
            10\%   AttrMask    & 72.87 ± 2.08 & 70.72 ± 3.72 & 74.96 ± 3.46 \\
            \hline
        
        \label{tab-graphcl-aug-select-ablation}
        \end{tabular}
    }
    \end{minipage}
    
    \caption{Ablation Study of the Aug Ratio of GraphCL Augmentations.}
    \begin{minipage}{1\linewidth}
    \centering
    \resizebox{1\hsize}{!}{
        \begin{tabular}{cccccc}
            \hline
            \multirow{2}{*}{Dataset}& Aug   & Node      & Edge          & \multirow{2}{*}{Subgraph} & Attribute \\
                                    & Ratio & Dropping  & perturbation  &                           & Masking   \\
            \hline
            NCI1    & 0.0       & 74.48 ± 1.91 & 74.48 ± 1.91    & 74.48 ± 1.91 & 74.48 ± 1.91 \\
            NCI1    & 0.1       & 75.01 ± 2.91 & 72.94 ± 2.41    & 74.77 ± 2.78 & 75.96 ± 2.28 \\
            NCI1    & 0.2       & 74.40 ± 2.84 & 72.07 ± 2.74    & 75.09 ± 2.46 & 75.57 ± 2.34 \\
            NCI1    & 0.3       & 74.57 ± 2.14 & 71.87 ± 2.14    & 74.18 ± 2.94 & 75.11 ± 2.24 \\
            NCI1    & 0.4       & 73.94 ± 2.32 & 70.29 ± 2.08    & 74.31 ± 2.48 & 75.13 ± 2.52 \\
            NCI1    & 0.5       & 73.70 ± 2.43 & 71.44 ± 2.24    & 74.55 ± 1.90 & 74.70 ± 2.01 \\
            \hline
        \end{tabular}
    }
    \end{minipage}
    \label{tab-graphcl-aug-ratio-ablation}
\end{table}



\subsection{The Effectiveness of Our Learnable Graph Augmentations}
\label{sec-mnist-exp}

\begin{figure*}
    \begin{center}
    \includegraphics[width=1\textwidth]{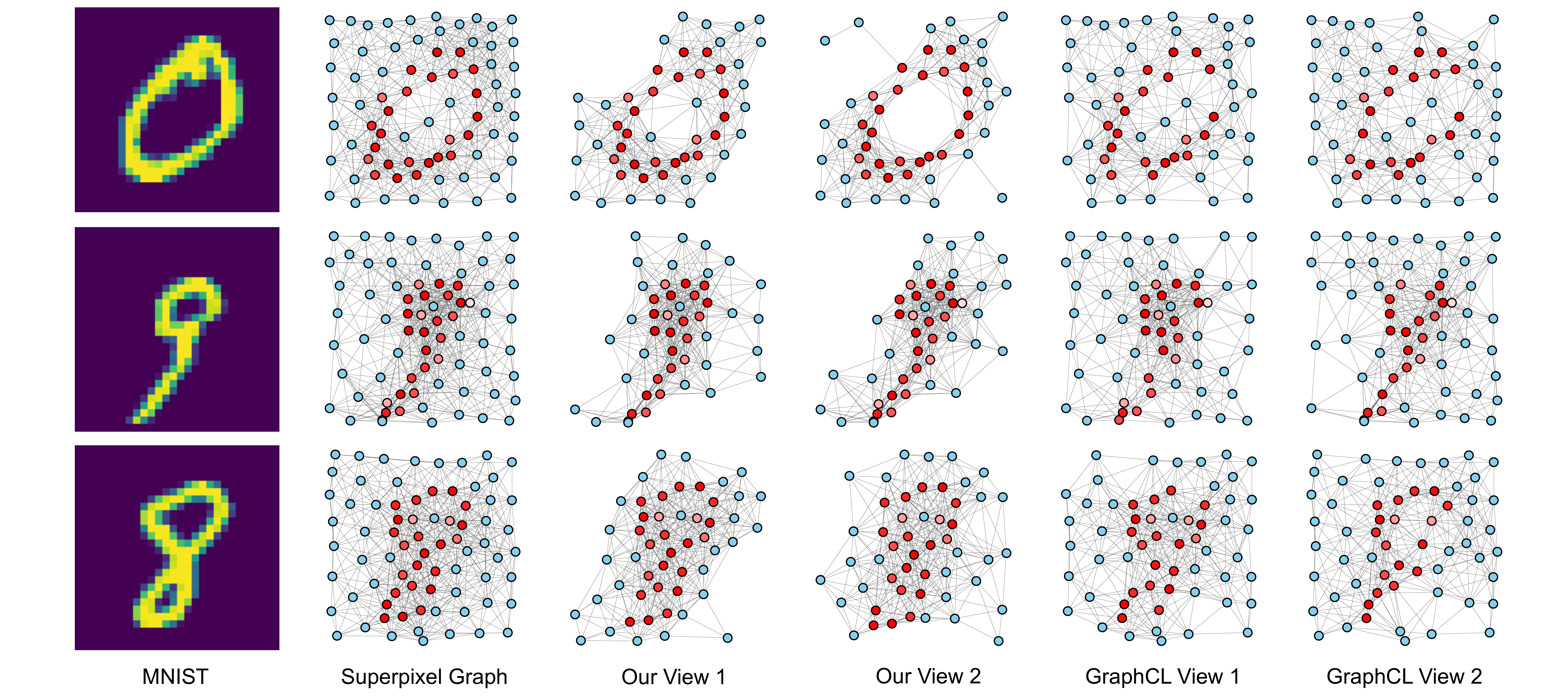}
    \end{center}
    \caption{View visualization on the MNIST-Superpixel dataset. The nodes with non-zero node attribute are colored in red, other nodes are colored in blue. Redder nodes indicate larger value of the node attribute.}
    \vspace{0.6cm}
    \label{fig-mnist-superpixel-vis}
\end{figure*}

\begin{figure*}
\begin{center}
    \includegraphics[width=1\textwidth]{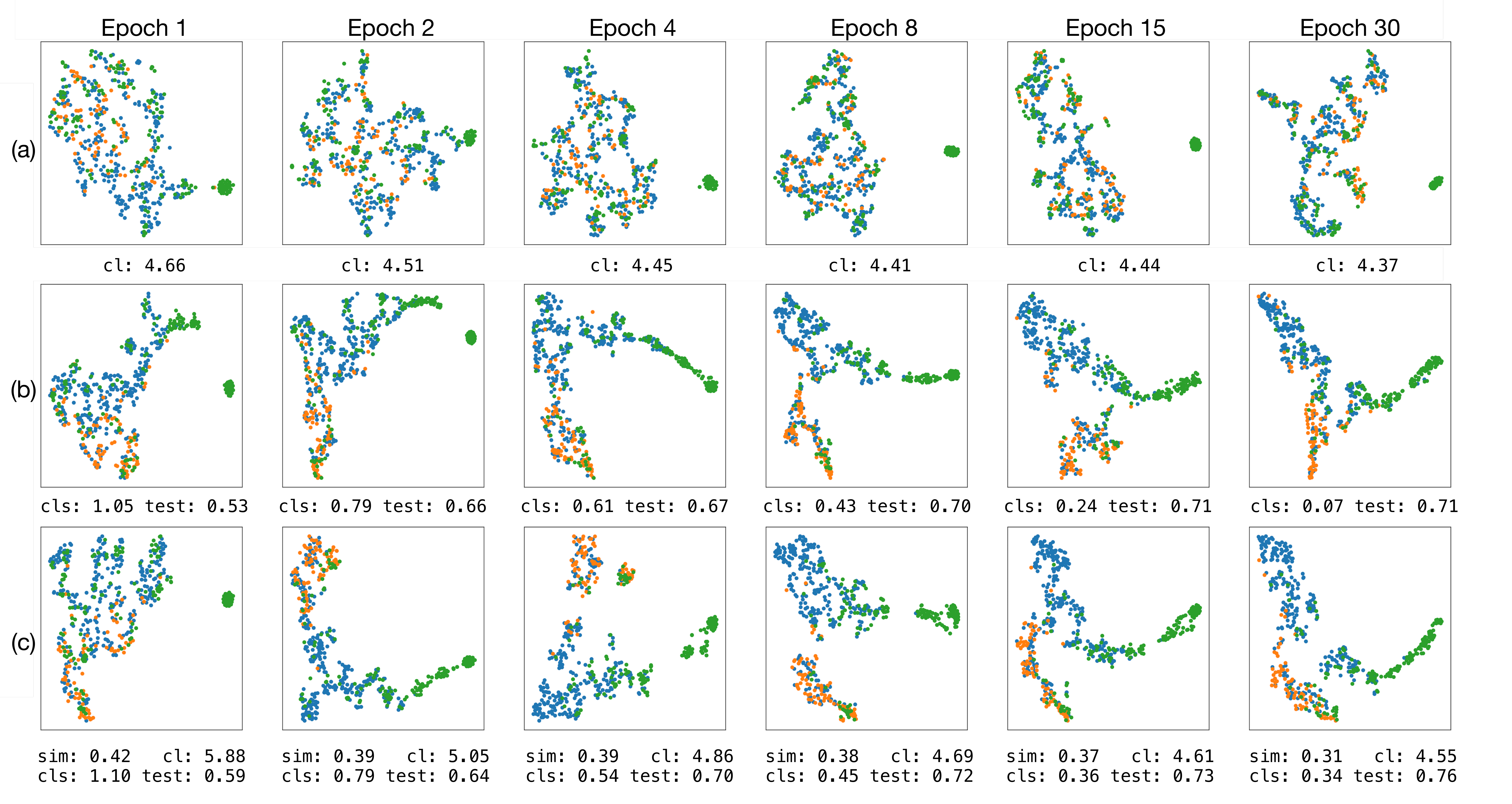}
\end{center}
\caption{t-SNE visualization of our naive-strategy and joint-strategy. (a) is the contrastive pre-training stage of naive-strategy. (b) is the fine-tuning stage of the native strategy. (c) is the training stage of joint-strategy. cl, cls, sim represent $\mathcal{L}_\text{cl}$, $\mathcal{L}_\text{cls}$, $\mathcal{L}_\text{sim}$, test represents the test accuracy.}
\vspace{0.6cm}
\label{fig-collab-tsne}
\end{figure*}

Here we demonstrate the superiority of learnable graph augmentation policies over the fixed ones. Since the graph datasets are usually difficult to be manually classified and visualized, we trained a view generator on MNIST-Superpixel Dataset \cite{monti2017mnistsuperpix} to verify that our graph view generator is able to effectively capture the semantic information in graphs than GraphCL \cite{you2020graphcl}. The visualization result is shown in Figure \ref{fig-mnist-superpixel-vis}.

The MNIST-Superpixel Dataset \cite{monti2017mnistsuperpix} is made of the super-pixel graphs of the MNIST Dataset \cite{lecun1998lenet}, contains 60000 training samples and 10000 testing samples, each graph have 75 nodes. The node attribute can be understand as the intensity of each super-pixel.

Here we jointly trained the view generators with the classifier until the test accuracy (evaluated on generated views) reached $90\%$. Since our only topological augmentation is node dropping. So we compared the view of Graphic's node dropping augmentation, and use the default setting $\text{aug\_ratio}=2$. Figure \ref{fig-mnist-superpixel-vis} shows that, our view generator are more likely to keep key nodes in the original graph, preserving its semantic feature, yet providing enough variance for contrastive learning. 


\newpage

\subsection{Analysis for Joint Training Strategy}
\label{sec-strategy-albation}

Here we compared the naive-strategy (Algorithm 1 in the paper) with the joint-strategy (Algorithm 2 in the paper). We trained on COLLAB \cite{collab} dataset, which have 5000 social network graphs of 3 classes, the average nodes and edges are 74.49 and 2457.78. Here we use 5-layer GIN \cite{xu2018gin} as the backbone for both the view generator and the classifier. For naive-strategy, there is 30 epochs of contrastive pretrain using 80\% unlabeled data and 30\% of fine-tuning using 10\% of data. For joint-strategy, there is 30 epochs of joint training. 

\begin{figure}[t]
    \begin{center}
        \includegraphics[width=1\linewidth]{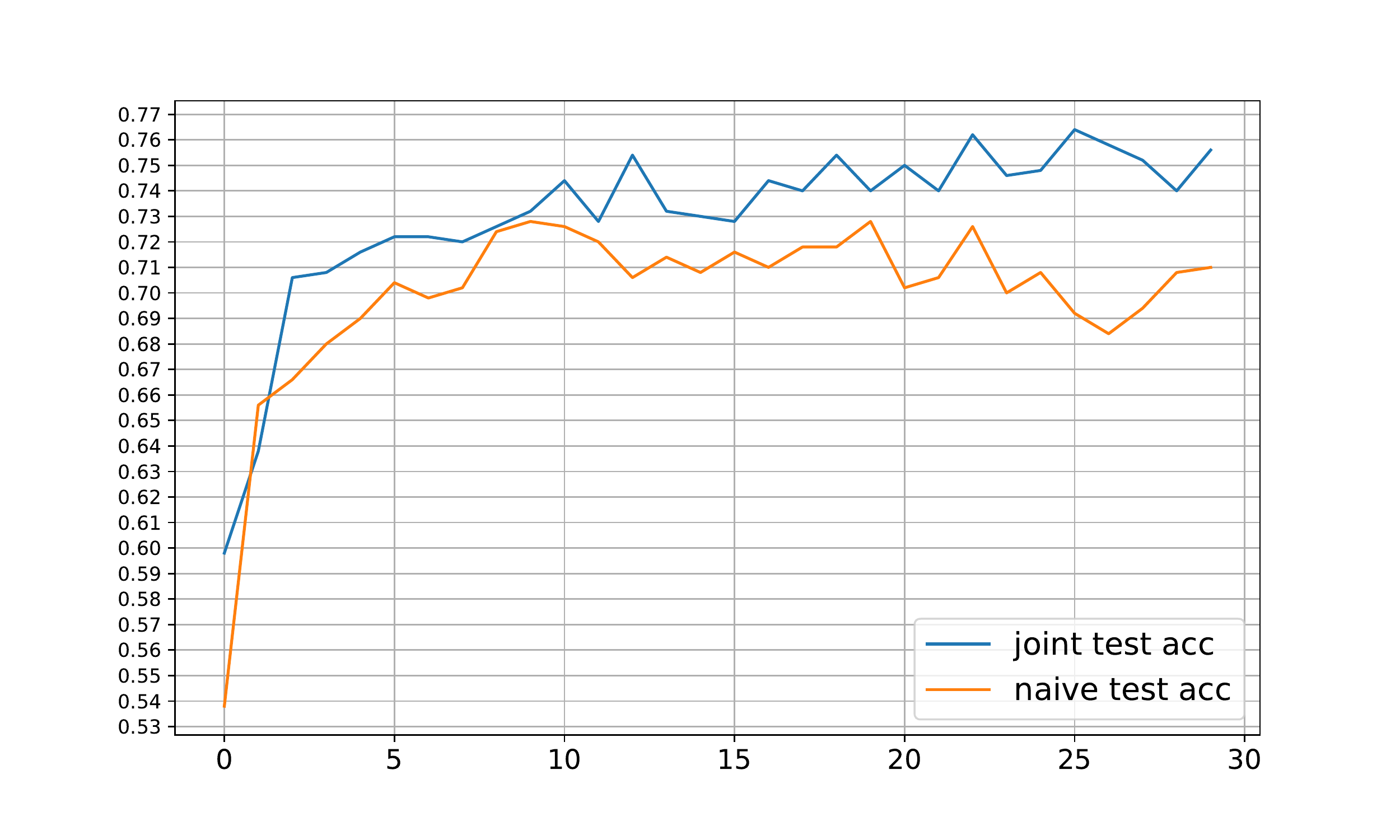}
    \end{center}
    \caption{Accuracy comparison between the naive-strategy and the joint-strategy.}
    \label{fig-acc-compare}
\end{figure}

We compared the learning curves in Figure \ref{fig-acc-compare}. The contrastive losses are both multiplied by $0.14$ to fit in the figure. Here we can see the $\mathcal{L}_\text{cls}$ of naive strategy drops much faster than the joint strategy. However, the test accuracy of naive strategy is lower than the joint strategy, and shows an downward tendency, indicating over-fitting. The joint strategy considerably alleviate the over-fitting effect, this also shows the effectiveness of our label-preserving view generator. 

We also visualize the process for learning the embedding for each strategy using t-SNE \cite{van2008tsne} in Figure \ref{fig-collab-tsne}. Figure \ref{fig-collab-tsne} (a) demonstrates that during the contrastive learning process, the graphs that have the same semantic label could gradually cluster together, but it still difficult to recognize the decision boundary to classify the graphs, while using labeled data to fine-tune the model (see figure \ref{fig-collab-tsne} (b)) could obtain much better graph representations for classification, indicating that to some extent, only using contrastive learning could benefit classification, but still far away from supervised learning. Figure \ref{fig-collab-tsne} (c) presents the joint training process, we can easily find that introducing label supervision, the model could learn better representations using a few epochs and looking at the sim, and cl loss values, both of them decrease, indicating that the views of one input graph are more different but the representations of these views are close enough, hence the view generators learn to generate different views and  preserve the semantic label of the input graph.

\begin{figure}[t]
    \begin{center}
        \includegraphics[width=1\linewidth]{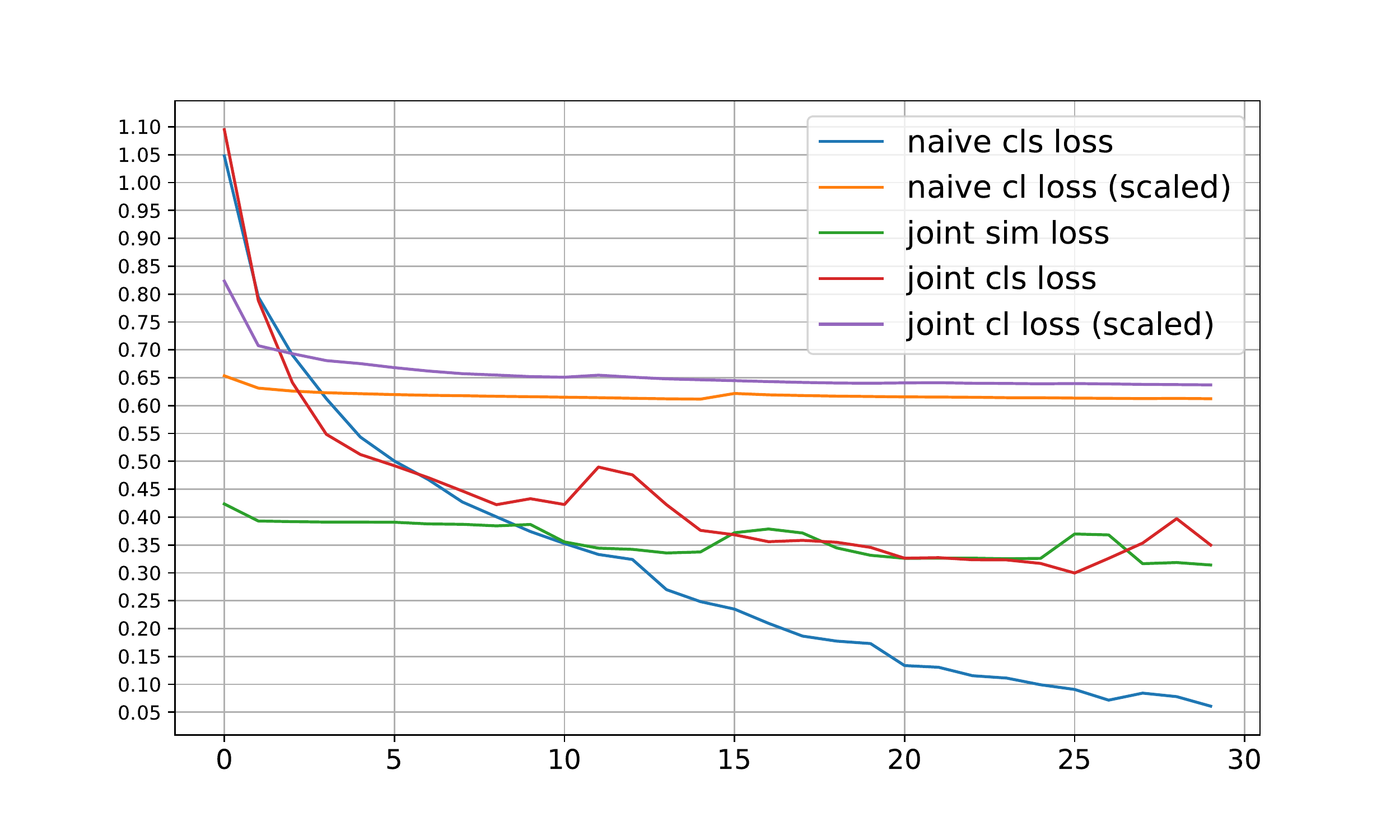}
    \end{center}
    \caption{Loss comparison between naive-strategy and the joint-strategy.}
    \label{fig-loss-compare}
\end{figure}



\section*{Acknowledgement}
We appreciate the efforts made by AC, SPCs, PCs and Reviewers for improving the manuscript. This work was done in part, when Mr. Yihang Yin did his research internship at Baidu Research. Q. Wang and H. Xiong were supported in part by National Key R\&D Program of China (No. 2018YFB1402600).

\newpage
\bibliography{aaai22}
\end{document}